# IMVB7t: A Multi-Modal Model for Food Preferences based on Artificially Produced Traits

Mushfiqur Rahman Abir*
20-42738-1@student.aiub.edu American International University-Bangladesh
Dhaka, Bangladesh

Md. Tanzib Hosain*
20-42737-1@student.aiub.edu American International University-Bangladesh
Dhaka, Bangladesh

Md. Abdullah-Al-Jubair†
American International University-Bangladesh Dhaka, Bangladesh abdullah@aiub.edu

M. F. Mridha‡
American International University-Bangladesh
Dhaka, Bangladesh
firoz.mridha@aiub.edu

## ABSTRACT

Human behavior and interactions are profoundly influenced by visual stimuli present in their surroundings. This influence extends to various aspects of life, notably food consumption and selection. In our study, we employed various models to extract different attributes from the environmental images. Specifically, we identify five key attributes and employ an ensemble model IMVB7 based on five distinct models for some of their detection resulted 0.85 mark. In addition, we conducted surveys to discern patterns in food preferences in response to visual stimuli. Leveraging the insights gleaned from these surveys, we formulate recommendations using decision tree for dishes based on the amalgamation of identified attributes resulted IMVB7t 0.96 mark. This study serves as a foundational step, paving the way for further exploration of this interdisciplinary domain.

## KEYWORDS

artificial intelligence, computer vision, food recommendation, intelligent system

## 1 INTRODUCTION

Food is an essential requirement for human sustenance [8]. Varied environments evoke distinct culinary preferences in individuals, with choices often influenced by both the setting and mood. Environments span a spectrum, from bustling restaurants to serene parks and tranquil beaches, while moods range from jubilant to melancholic and, angry to placid [11]. This confluence of place and emotion often leads to specific cravings. For instance, one might yearn for ice cream while basking on a sun-kissed beach in elation, or crave a hearty burger amidst the quiet solitude of a park in sorrow. Meanwhile, a neutral demeanor in the familiar confines of a restaurant might prompt a preference for crisp salads [12].

Entering the food suggestion system, a sophisticated solution is poised to revolutionize dining experiences by offering tailored dish recommendations based on both environmental and emotional states [13]. Leveraging the capabilities of computer vision to decipher environmental cues and the analytical process of data mining to discern patterns and preferences, this system promises to cater to individual tastes with unprecedented precision and insight [7].

We propose a model, specifically used a variety of models to extract distinct features from the photos of the surroundings. In particular, for some of their detection, we select five critical qualities and use an ensemble model, IMVB7, built on five unique models. In order to identify trends in dietary choices in response to visual stimuli, we also carried out questionnaires. Using the information obtained from these surveys, we create dish recommendations by combining the identified attributes.

The subsequent sections of this paper delve into the intricacies of the proposed method, providing comprehensive details on each aspect. Specific attention is paid to the datasets utilized, convolutional neural network (CNN) models employed, and other pertinent modules, each discussed in dedicated subsections. Following the presentation of the findings, the following sections offer an analysis and discussion of the findings, elucidating the implications and significance of the research results. In addition, potential avenues for future research in this domain are explored, paving the way for further innovation and advancement.

## 2 MOTIVATION

Although previous research has delved into the detection and identification of various scene elements from images, such as weather conditions, and implementations of data mining have offered food suggestions based on historical data, a critical gap remains: the direct derivation of comprehensive food recommendations from environmental imagery. This void presents a compelling opportunity, holding immense potential for both human welfare and food industry. Motivated by the pressing need for a solution that seamlessly integrates visual stimuli with gastronomic recommendations, our research aims to introduce a robust and versatile suggestion system. Envisaged applications span diverse settings, from restaurants and cafés seeking to enhance customer experiences, to food delivery services aiming to personalize offerings based on location, and food recommendation platforms tailoring suggestions to individual moods. Grounded in the extraction of attributes from given images, our approach represents a novel and impactful step toward culinary innovation and consumer satisfaction.

## 3 RELATED WORKS

In recent years, the proliferation of data science and computer vision has inspired many researchers to contribute to the idea of food recommendation system using computer vision. In 2023, Zhang *et al.* worked on a food recommendation method using temporal-dependent graph neural networks and data augmentation, which results in accurate and robust recommendations [18]. In the same year, Ajami *et al.* contributed to a project about a food recommender system in academic environments [1]. His AdaBoost model showed 73.7% accuracy. Tran *et al*. compared VGG16, MobileNet, ANN, Resnet18, Resnet50, Densenet121 and many more models for image recognition [17]. The relationship between food, lifestyle, generics, and disease is deeply analyzed by Zhang *et al.* [18]. In 2021, Dhali *et al.* proposed a paper which largely covered different approaches to estimate calories from food images, discussing the complexities, and suggesting simpler methods for end-users [4]. Chiu *et al.* introduced a method for location-based food recommendations in 2022 [3]. Min *et al* [10] detailed the framework of food recommendations and challenges. The unified framework proposed in this paper for food recommendation considers context, domain knowledge, and personal models. Ge *et al.* health-aware recommender system provided the logic to build a personalized health-aware system [2]. Axelson *et al.* showed the impact of culture and food-related behaviors [9]. D. Liu *et al.* stated that weather plays a significant role in shaping online food ordering habits, impacting both the types of food ordered and the frequency of orders. Seasonal changes often influence culinary preferences, with lighter fares like salads and smoothies being favored during hot summer months, while heartier dishes such as soups and stews are more popular in colder weather [9]. Diet-right architecture provided a person's diet and health concerns [5].

## 4 METHODOLOGY

The aim of this study lies a sophisticated CNN-based image classification system tailored to our food recommendation framework. Within this framework, five distinct CNN-based image classification models, in conjunction with a bespoke algorithm designed to analyze five key attributes, collaborate to generate five unique outputs for a given image.

In each iteration, the output of each model was transformed into a relevant binary vector that encapsulates the presence or absence of specific attributes. These vectors were then appended to an initial empty vector, gradually aggregating attribute information across iterations. Upon completion of predictions for all attributes, the concatenated vector undergoes reshaping to align with the input requirements of the decision tree classifier model.

Subsequently, the reshaped vector is fed into the classifier to predict the recommended culinary genre for the input image. The training of the five CNN models involved the utilization of diverse architectures and datasets. Given the constraints posed by computational resources, emphasis was placed on striking a balance between model accuracy and architectural simplicity.

To address these challenges, "Transfer Learning" techniques were employed, leveraging models with pre-trained weights from TensorFlow. Additionally, ensemble model architectures were developed where they demonstrated superior performance in terms of accuracy and loss compared to pre-trained alternatives. For the final label classification, a decision tree classifier was used for the final label classsification.

The overarching system workflow is depicted in Figure 1.

### 4.1 Datasets

Training our 5 pivotal models with proper and adequate data was a critical task for this project. We decided to create our own dataset using generative artificial intelligence, as we are trying to recommend food by analysing generated images. We selected 5 main attributes to detect and extract information from an environmental image to recommend suitable foods, and those are scene, dominant color, weather, and period. We also implemented age detection for a more specific result for a person, as people of different ages tend to like different kinds of cuisine. For the image generation, we have used Stable Diffusion and DALL-E mainly.

We divided the image generation and the image data into 5 different classes matching out 5 attributes. Each class comprises 10000 images totaling 50000 images for the whole dataset, shown in Figure 2. This substantial dataset size allowed for sufficient variation and diversity within each attribute category, thereby enhancing the generalizability and effectiveness of our models.

In image generation, the generation prompt playes a very crucial role. We have very cautiously engineered the prompts to get right image outputs. For example, *A rural countryside scene with rolling hills, farmland, and wind turbines generating clean energy.* We tried to make the environmental image as detailed as possible using specific prompts. Furthermore, to ensure the integrity of our evaluation process, we adopted a rigorous approach to dataset splitting. The dataset was randomly partitioned into training, validation, and testing sets, maintaining a consistent distribution of images across all classes. Specifically, 80 percent of the images were allocated to the training set, 10 percent to the validation set, and the remaining 10 percent to the testing set. This partitioning scheme helped mitigate bias and overfitting while enabling robust model evaluation on unseen data.

### 4.2 Deep learning models

Rather than relying on existing models and architectures, we implemented an ensemble CNN model (IMVB7) tailored for our use case. We needed to adequately capture the cues of an environment for suggesting appropriate food. We also integrated multiple models, each specializing in capturing and analyzing different aspects of an event, into the ensemble model for better results. We have used transfer learning for further improvements to our model, allowing us to leverage pre-trained models on large datasets to extract features relevant to the food recommendation task. Figure 3 represents the full architecture of the proposed ensemble model. There are $n$ iteration of the ensemble model. An iteration starts with the Self-pace Factor (SPF) calculation block. Then it passes to the Current Ensemble $F(x)$ block. After that self-paced under-sample mechanism handles the data and then passed it to the next block which is named balance training set. Balance training set is divided into majority set and minority set. The majority set processes the passed data. Now in case of the $i^{th}$ iteration the final stage data passes through a set of majority bins named Bin 1, 2, 3 ... and so on. Also the final data are passed to the mentioned 4 different models for further processing and then to root recognizer which help in recognizing the different data streams. And then the organized data is given down to the training block. Training block perfectly trains the data and then again sends the data to a new root recognizer. From here the self-pace factor and current ensemble $F(x)$ are transformed to the final ensemble recognizer IMVB7.

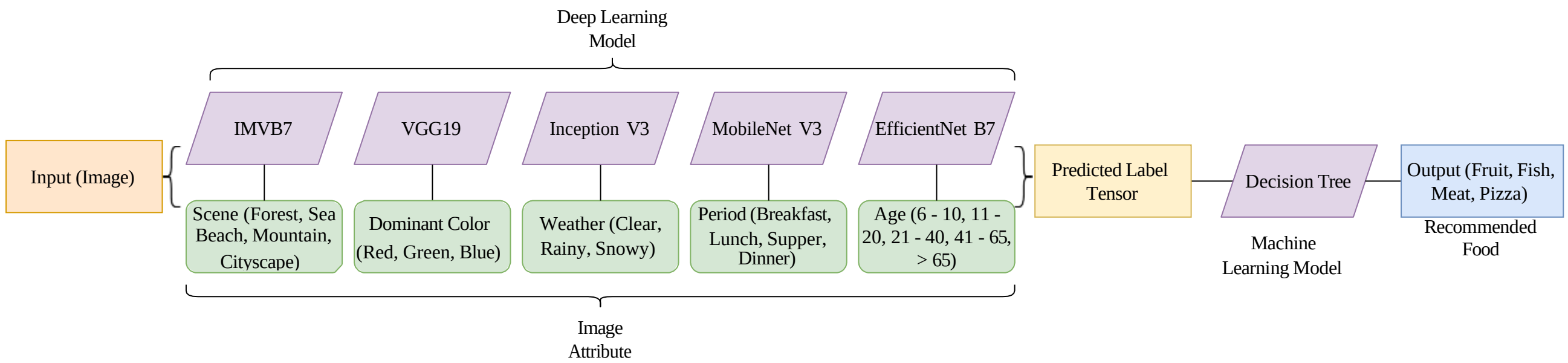

**Figure 1: System workflow of the full project.**

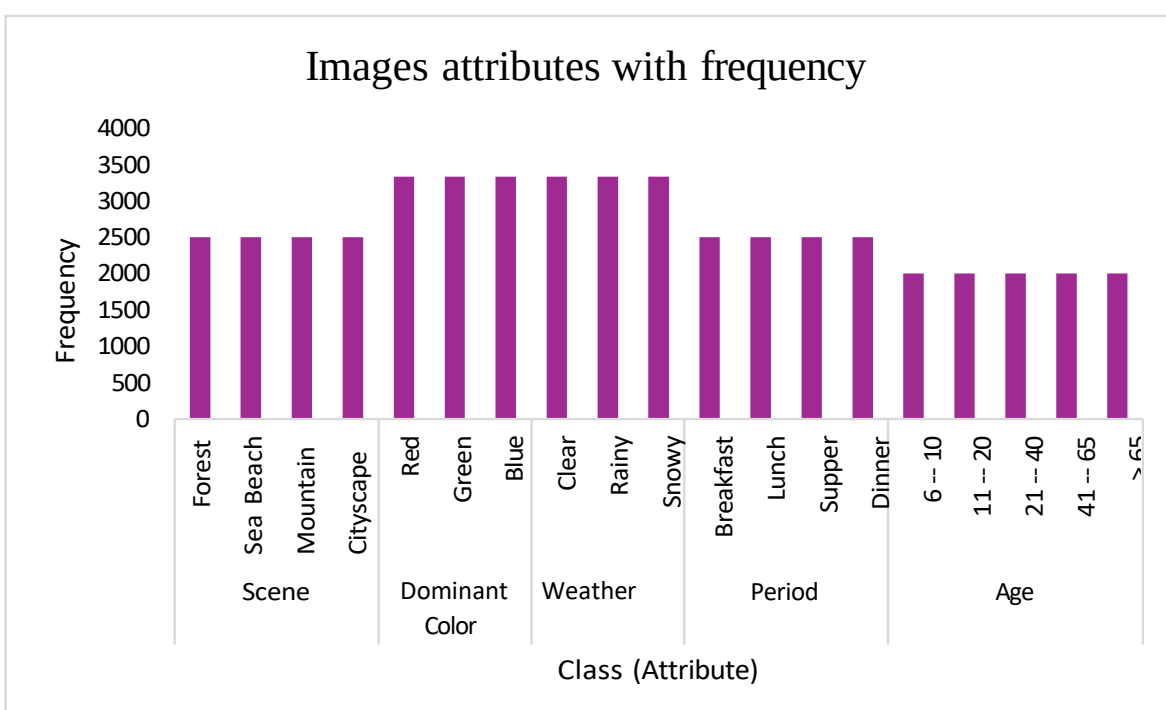

**Figure 2: Images attributes with their respective count.**

In addition to the ensemble model mentioned above, we added 4 CNN models to identify the dominant color, weather condition, period of day, and age of the input image which is visualized in Figure 1 diagram. From the system workflow diagram we can see that we used a machine learning decision tree model to process the final data right before the output.

Finally, the models are each specifically good for specific tasks: For scene attribute detection: IMVB7, dominant color detection: VGG19 [14], weather condition detection: Inception V3 [15], period of day detection: MobileNet V3 [6], and EfficientNet B7 [16] for age detection.

## 4.3 Survey

In our research, we conducted a comprehensive survey to gain insight into the preferences and tastes of individuals about food choices within various environmental contexts. Understanding these preferences was essential to ensuring the effectiveness and relevance of our food recommendation system. With 4 distinct attributes utilized to identify each input image, 120 combinations of attributes were mathematically possible. However, not all combinations were considered practical or plausible.

After careful consideration and review, we narrowed down our focus to 120 viable combinations for inclusion in the survey. Each combination represented a unique environmental setting. The survey questions were designed to be straightforward, with participants presented with an image corresponding to one of the 75 attribute combinations and asked to select their preferred food choice from 4 options.

The participants were presented with 75 distinct images, each representing a unique combination of environmental attributes. For each image, participants were asked to select their preferred food choice from a set of 4 options. Before participating in the survey, all participants were provided with the necessary background information and their consent was obtained. This ensured that participants were fully informed and willing to participate in the survey process.

## 4.4 Machine learning classifier

In our study, we used a decision tree classifier to determine the final product, that is, the recommended food choice, based on the results of the five attributes. However, utilizing a decision tree classifier posed a challenge as our dataset consisted of categorical attributes. Since decision tree classifier do not yield accurate results when applied to categorical data directly, it was imperative to convert the data set into numeric values.

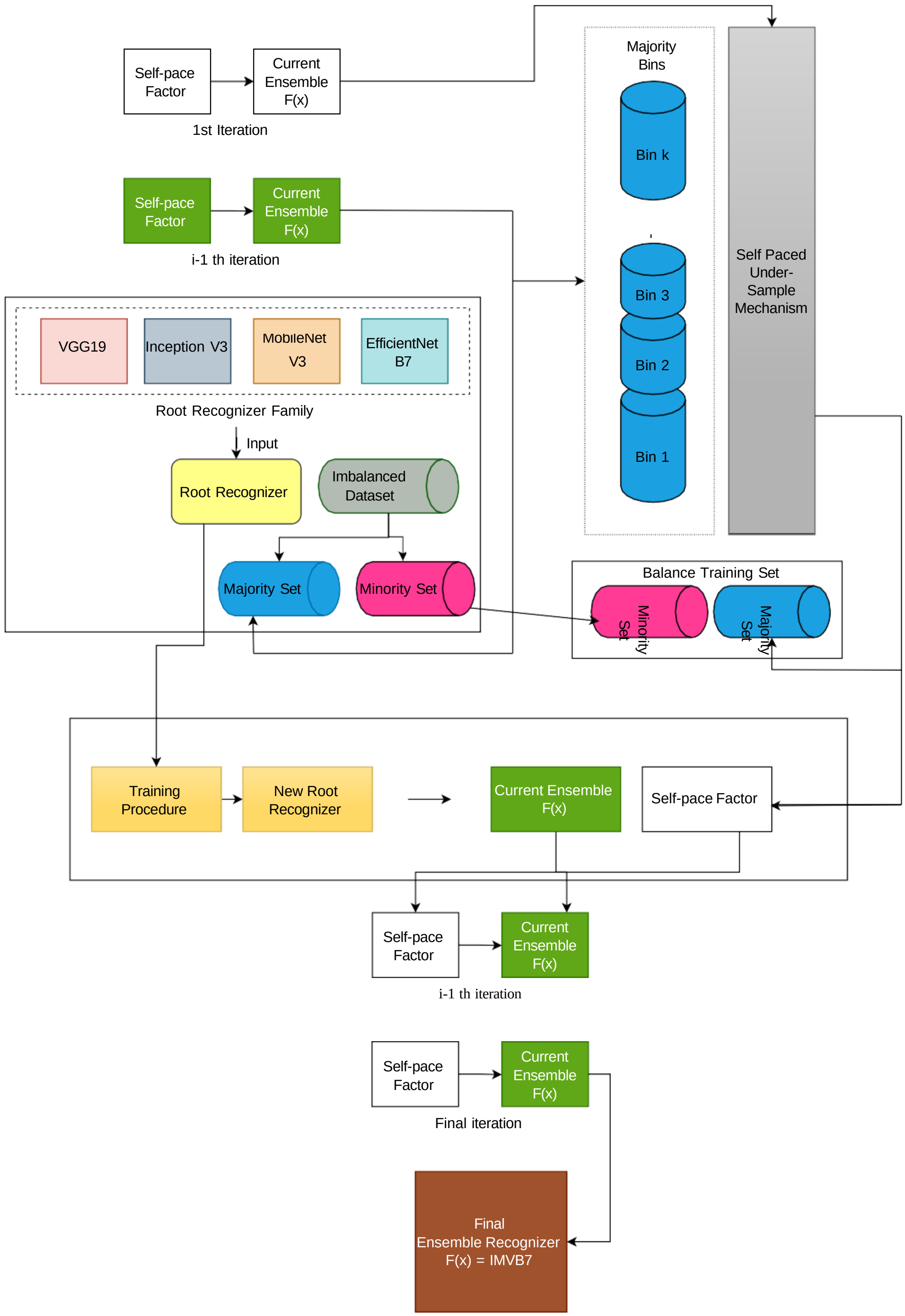

**Figure 3: Architecture and working process of IMVB7.**

To address this issue, we transformed the categorical attribute values into binary values. Subsequently, the values were taken into the tensor. This approach ensured that all attribute values maintained equal precedence, preventing the introduction of biases in the classification process.

Once the data set was properly converted, it was divided into 80 percent of the data allocated to the training set, 10 percent to the validation set, and the remaining 10 percent to the testing set. Subsequently, the dataset was fed into the decision tree classifier for training.

## 5 EVALUATION METRICS

In the intricate dance of classification analysis, we waltz through the delicate interplay between the classifier's decisions and ground truth. Initially, a multifaceted classification puzzle was presented itself, adorned with three distinct classes. However, with a twist of perspective, we unfurl four binary classification conundrums, each pondering whether an instance finds solace within a designated class.

In this binary ballet, our metrics accuracy, $A$ don elegant attire:

$$A = \frac{1}{|Te|} \sum_{x \in Te} I[\hat{c}(x) = c(x)] \quad (1)$$

Here, $\hat{c}(x)$ waltzes with the actual class $c(x)$, as dictated by the ground truth. $T_e$ becomes as our test set, whereas the elusive $I[\cdot]$ sweeps gracefully as the indicator function. Positive and negative classes pirouette as $\oplus$ and $\ominus$, respectively.

However, what is the grand ensemble of multiclass classifications? We invoke melodious strains of the macro-averaged F1 score:

$$F = \frac{(\beta^2 + 1)PR}{\beta^2 P + R} \quad (2)$$

where $\beta$ assumed the humble guise of 1. Precision, $P_{\text{macro}}$, and recall, $R_{\text{macro}}$, perform a duet of their own:

$$P_{\text{macro}} = \frac{1}{|C|} \sum_{i=1}^{|C|} \frac{TP_i}{TP_i + FP_i} \quad (3)$$

$$R_{\text{macro}} = \frac{1}{|C|} \sum_{i=1}^{|C|} \frac{TP_i}{TP_i + FN_i} \quad (4)$$

Where $C$ represents our ensemble of classes, and $TP_i$, $FP_i$, and $FN_i$ echo the trials and tribulations of true positives, false positives, and false negatives.

## 6 RESULTS ANALYSIS

Table 1 presents the performance metrics for various models across different tasks. The models evaluated include VGG19 for dominant color detection, Inception V3 for weather detection, MobileNet V3 for period of day detection, and EfficientNet B7 for age detection. Our models, IMVB7 and IMVB7t, were assessed for scene detection and food recommendation, respectively. The highest performance was achieved by our model (IMVB7t) in the food recommendation task, with an accuracy, precision, recall, and F1-score of 0.96. For scene detection, our model (IMVB7) also demonstrated superior performance with all metrics at 0.85 or higher. Among the pre-existing models, EfficientNet B7 showed the best performance for age detection with metrics of 0.84 or higher.

### 6.1 Convolutional neural network models

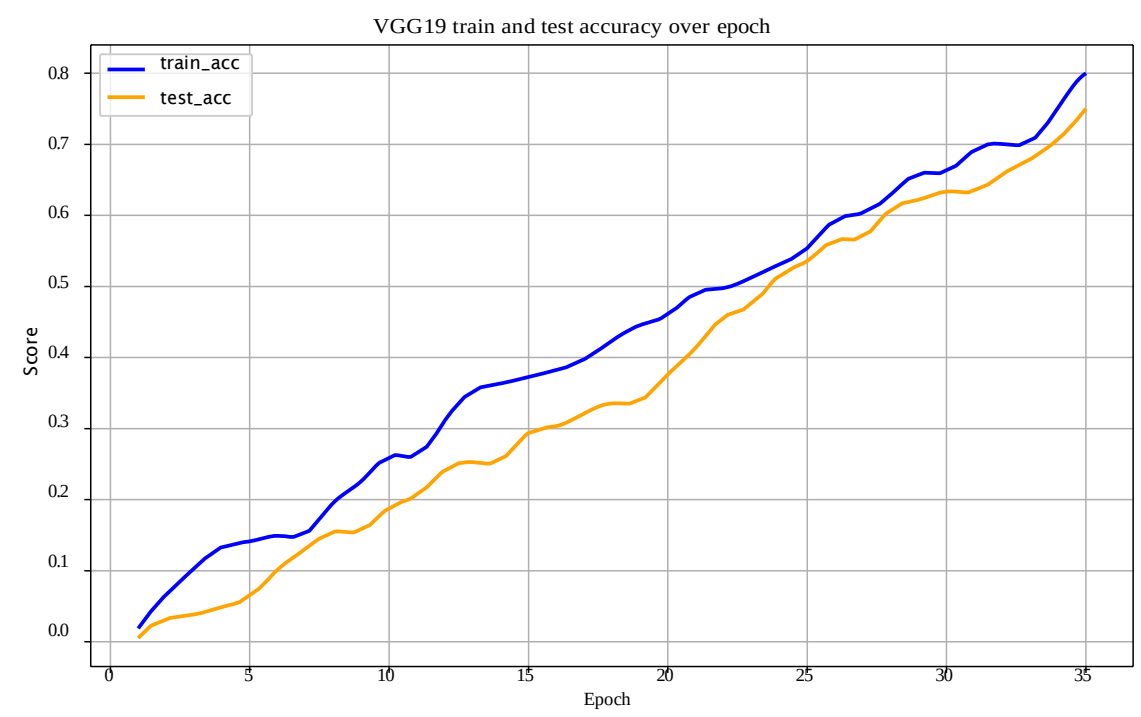

**Figure 4: VGG19 train and test accuracy over epochs.**

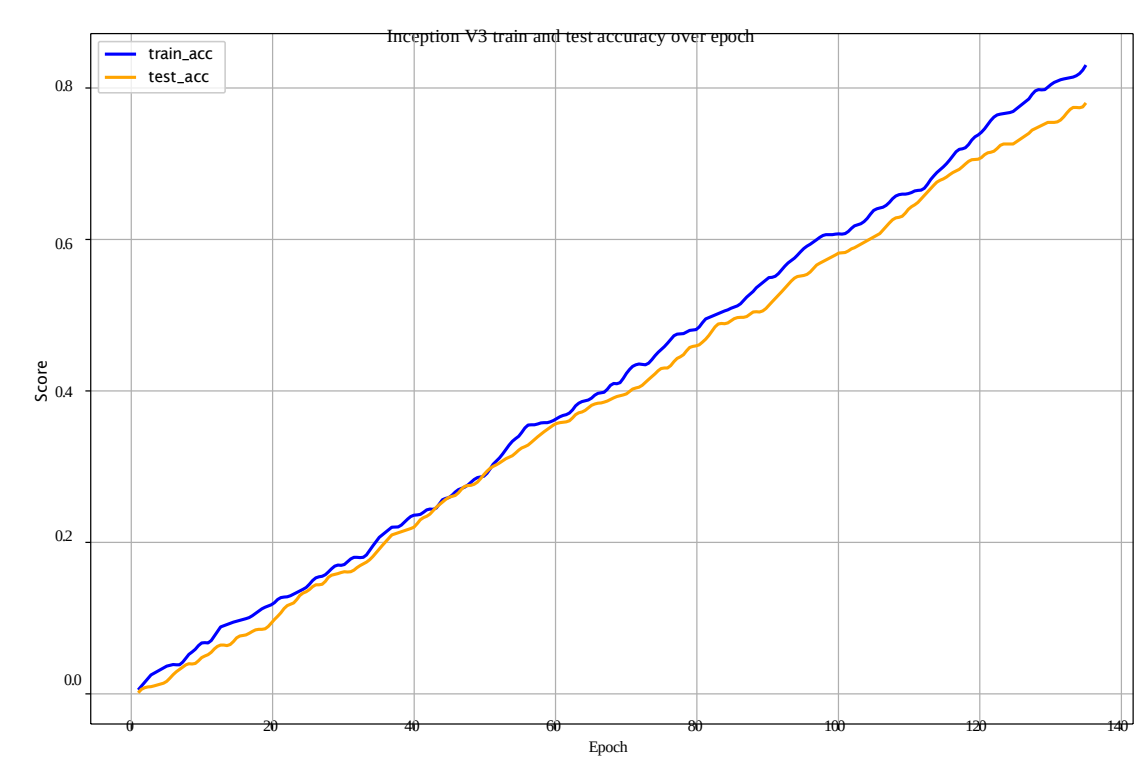

**Figure 5: Inception V3 train and test accuracy over epochs.**

The performance of our deep learning models in different image analysis tasks was assessed and depicted through validation and testing accuracies across multiple epochs. Figure 4 illustrates that VGG19, employed for dominant color detection, shows a consistent improvement in validation accuracy over 35 epochs, stabilizing the testing accuracy at approximately 0.75. The Inception V3 model in Figure 5, tasked with weather detection, demonstrates increasing validation accuracy over 135 epochs, with testing accuracy plateau- ing near 0.78 after an initial spike. Figure 6 shows that MobileNet V3, which is responsible for the period of day detection, displays improved validation accuracy over 45 epochs, while testing accuracy remains around 0.73 after an initial adjustment. EfficientNet B7, employed for age detection, depicts a continuous rise in validation accuracy throughout 375 epochs, with testing accuracy stabilizing around 0.84 after initial improvement, as illustrated in Figure 7. The ensemble model IMVB7, utilized for scene detection, exhibits a steady improvement in training accuracy, while testing accuracy fluctuates owing to noise and extraneous elements, eventually converging around 0.85 throughout 185 epochs, as shown in Figure 8.

**Table 1: Performance metrics for different models.**

| Model | Task | Accuracy | Precision | Recall | F1-Score |
|---|---|---|---|---|---|
| VGG19 | Dominant color detection | 0.75 | 0.76 | 0.75 | 0.75 |
| Inception V3 | Weather detection | 0.78 | 0.79 | 0.78 | 0.78 |
| MobileNet V3 | Period of day detection | 0.73 | 0.74 | 0.73 | 0.73 |
| EfficientNet B7 | Age detection | 0.84 | 0.85 | 0.84 | 0.84 |
| **Proposed Model (IMVB7)** | Scene detection | **0.85** | **0.86** | **0.85** | **0.85** |
| **Proposed Model (IMVB7t)** | Food recommendation | **0.96** | **0.96** | **0.96** | **0.96** |

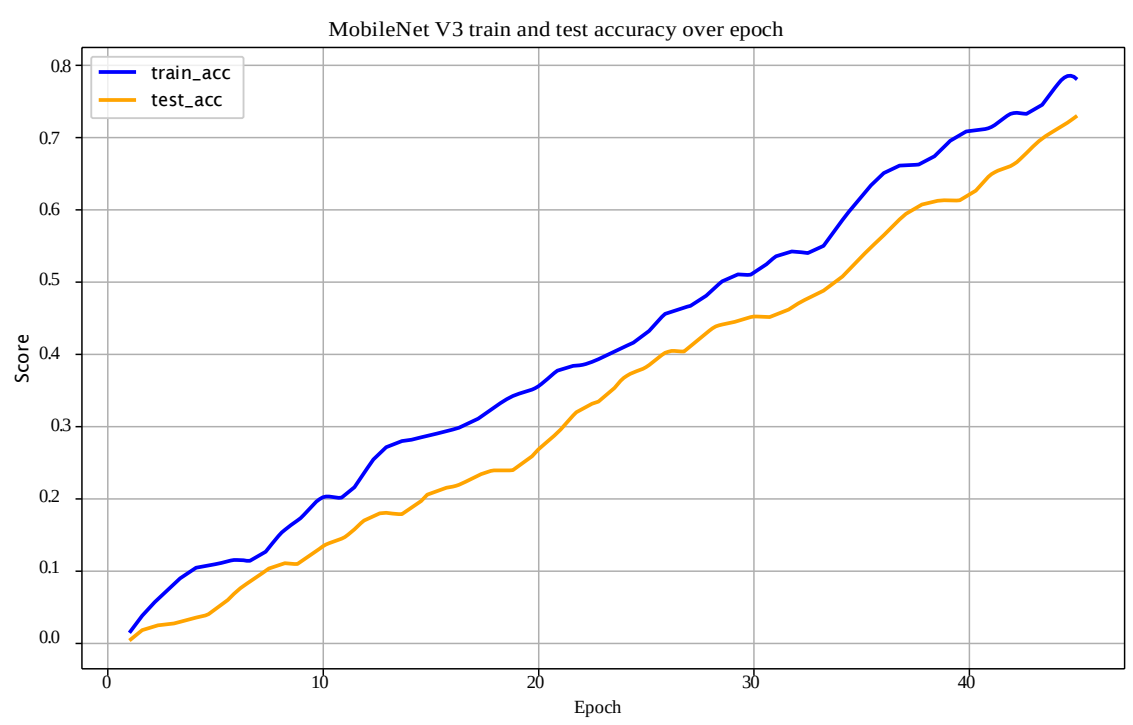

**Figure 6: MobileNet V3 train and test accuracy over epochs.**

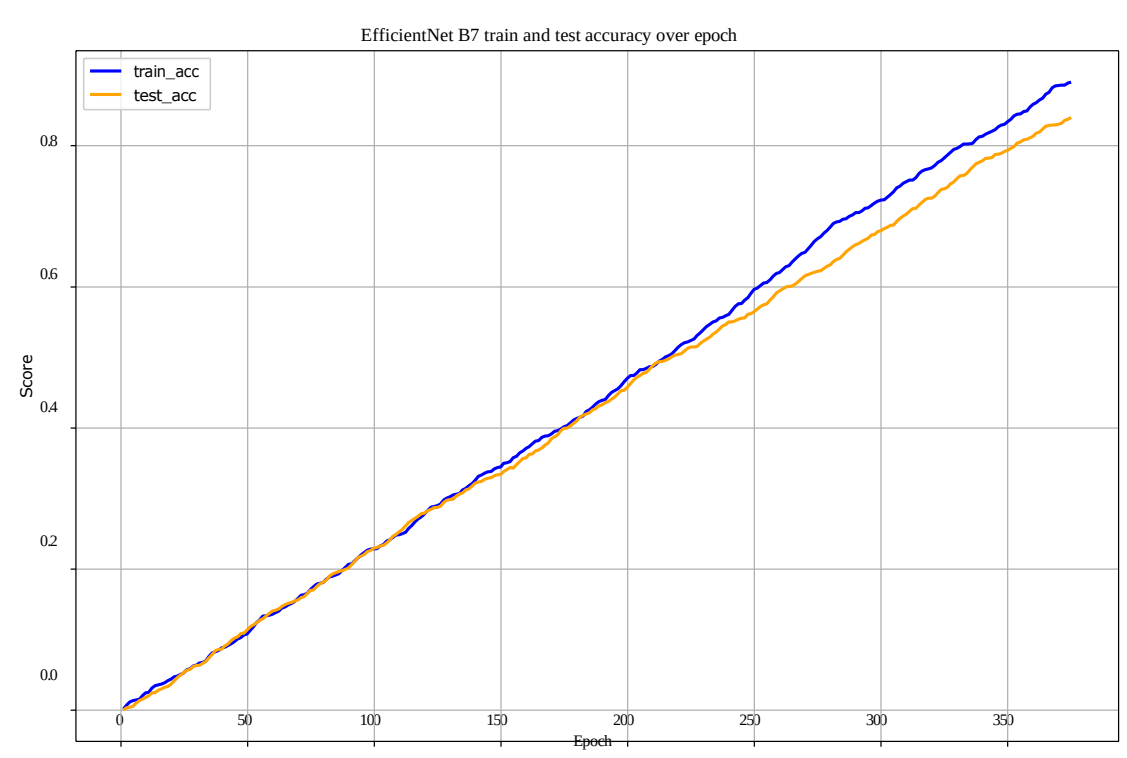

**Figure 7: EfficientNet B7 train and test accuracy over epochs.**

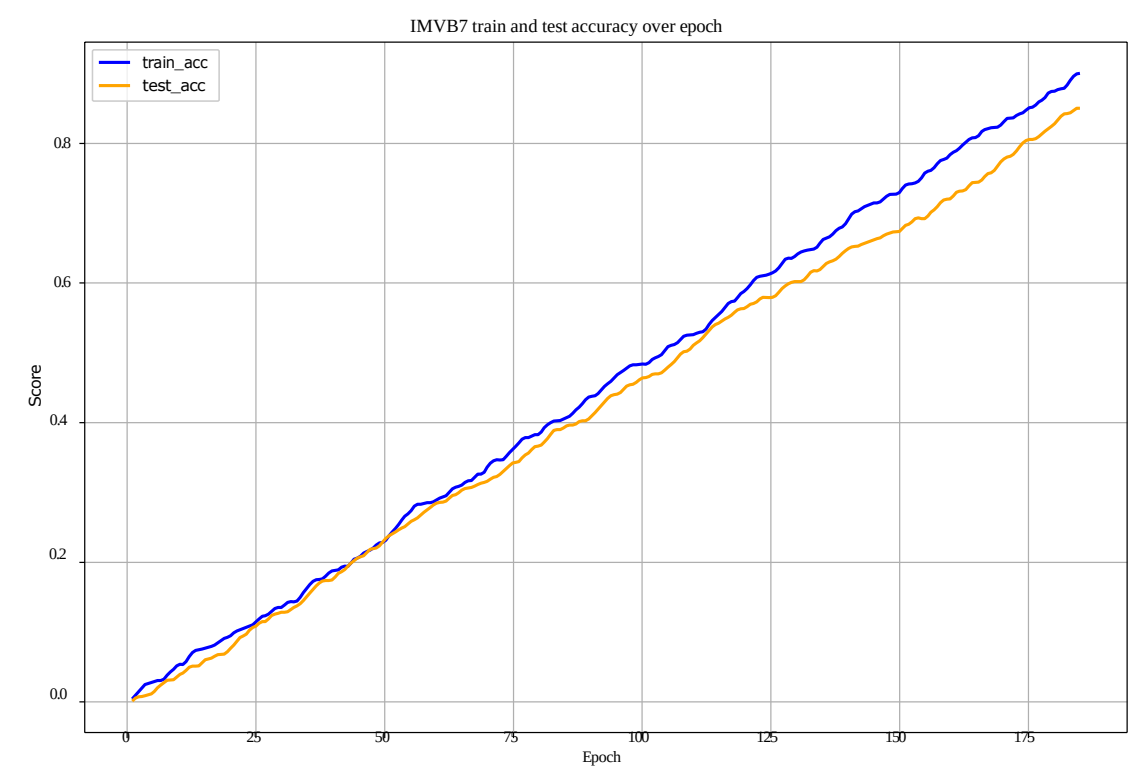

**Figure 8: IMVB7 train and test accuracy over epochs.**

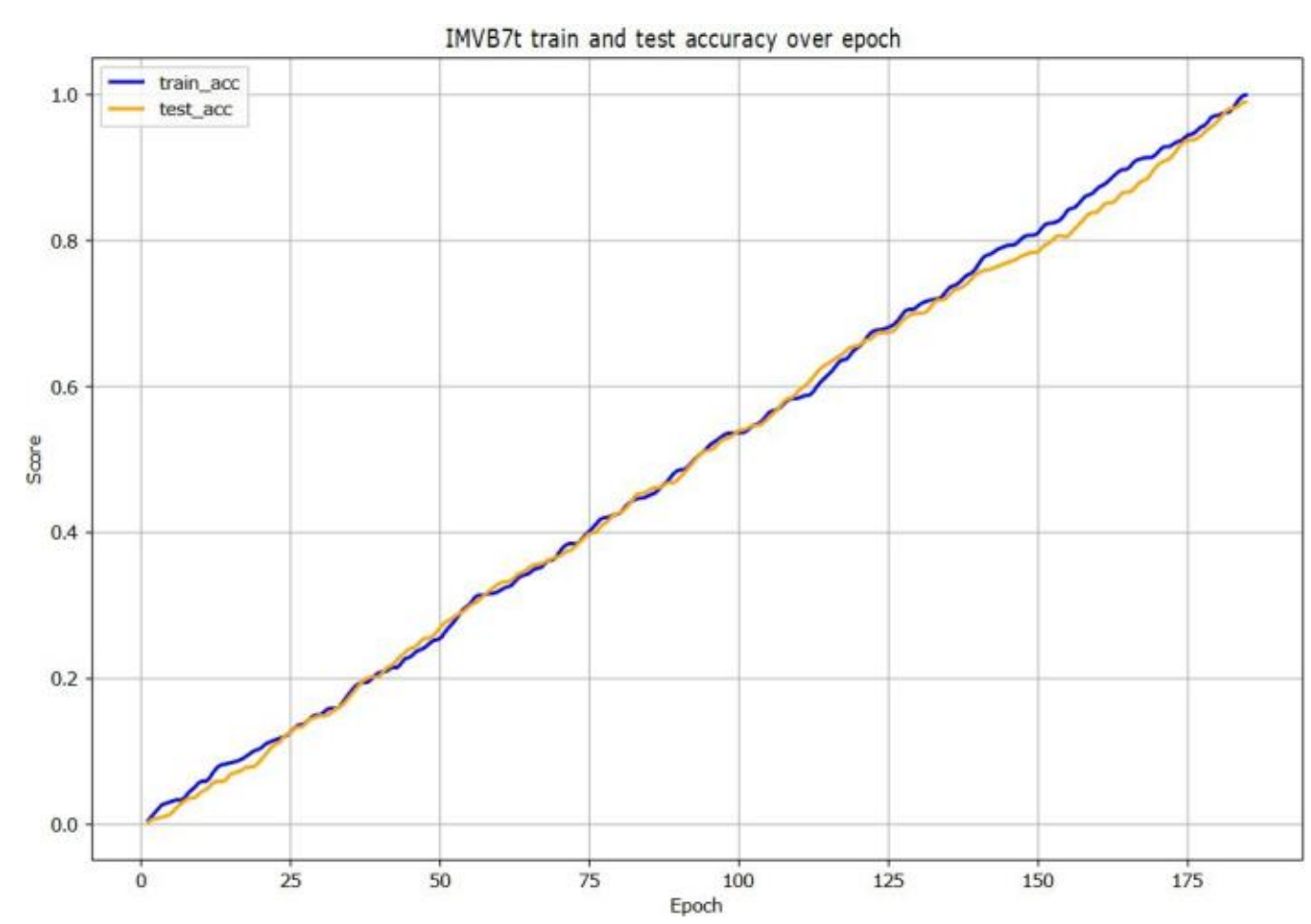

**Figure 9: IMVB7t train and test accuracy over epochs.**

## 6.2 Decision tree classifier

Figure 9 outlines the classification performance of the decision tree classifier utilized IMVB7t, predominantly high test accuracy values, often reaching the 0.96 mark for food suggestions, encompassing the Fruit, Fish, Meat, and Pizza output classes. A total of the 75 instances were analyzed.

## 7 FUTURE DIRECTIONS AND CONCLUSION

Expanding the scope of this study to encompass a broader range of environmental features and their influence on food preferences could yield valuable insights. Moreover, the commercial potential of our system in the food industry merits exploration, with opportunities for integration into food delivery services, restaurant recommendation platforms, and personalized meal planning applications. This study contributes to the intersection of machine learning and culinary arts. Although machine learning has seen significant advancements in various domains such as healthcare, agriculture, and entertainment, its application in food recommendations based on environmental cues is relatively unexplored. Leveraging deep learning and data mining techniques, our work pioneers a novel approach to food recommendation, paving the way for future re- search in this domain. Furthermore, our system has the potential to improve user experience in food-related platforms and services.

## ACKNOWLEDGEMENTS

Part of this work's sections, including text, tables, and figures was prepared by taking the advantage of GPT-4o.